\documentclass[10pt,twocolumn,letterpaper]{article}

\usepackage{cvpr}
\usepackage{times}
\usepackage{epsfig}
\usepackage{graphicx}
\usepackage{amsmath}
\usepackage{amssymb}

\usepackage{comment}
\usepackage{verbatim}
\usepackage{color}
\usepackage{xspace}

\usepackage{caption}
\usepackage{subfigure}
\usepackage{footnote}
\makesavenoteenv{tabular}
\makesavenoteenv{table}

\makeatletter
\DeclareRobustCommand\onedot{\futurelet\@let@token\@onedot}
\def\@onedot{\ifx\@let@token.\else.\null\fi\xspace}

\def\eg{\emph{e.g}\onedot} 
\def\ie{\emph{i.e}\onedot} 
 
\def\etc{\emph{etc}\onedot}

\makeatother

\newcommand{\DoubleFigureWidth}{230pt}

\newcommand{\SingleFigureWidth}{200pt}
\newcommand{\SmallSingleFigureWidth}{160pt}

\newcommand{\FifthFigureWidth}{96pt}

\usepackage[pagebackref=true,breaklinks=true,letterpaper=true,colorlinks,bookmarks=false]{hyperref}

\cvprfinalcopy 

\def\cvprPaperID{413} 

\ifcvprfinal\fi

\begin{document}

\title{A Revisit on Deep Hashings for Large-scale Content Based Image Retrieval}

\author{Deng Cai\hspace{30pt}  Xiuye Gu \hspace{30pt} Chaoqi Wang\\
The State Key Lab of CAD\&CG\\
Zhejiang University\\
Hangzhou, P. R. China\\
{\tt\small (dcai,gxy0922,cqwong)@zju.edu.cn}
}

\maketitle

\begin{abstract}
	There is a growing trend in studying deep hashing methods for content-based image retrieval (CBIR), where hash functions and binary codes are learnt using deep convolutional neural networks and then the binary codes can be used to do approximate nearest neighbor (ANN) search. All the existing deep hashing papers report their methods' superior performance over the traditional hashing methods according to their experimental results. However, there are serious flaws in the evaluations of existing deep hashing papers: (1) The datasets they used are too small and simple to simulate the real CBIR situation. (2) They did not correctly include the search time in their evaluation criteria, while the search time is crucial in real CBIR systems. (3) The performance of some unsupervised hashing algorithms (\eg, LSH) can easily be boosted if one uses multiple hash tables, which is an important factor should be considered in the evaluation while most of the deep hashing papers failed to do so. 
	We re-evaluate several state-of-the-art deep hashing methods with a carefully designed experimental setting. Empirical results reveal that the performance of these deep hashing methods are inferior to multi-table IsoH, a very simple unsupervised hashing method. Thus, the conclusions in all the deep hashing papers should be carefully re-examined.
\end{abstract}

\section{Introduction} \label{sec:intro}

Content-based image retrieval (CBIR) \cite{CBIRsurvey} is an interesting and popular problem in computer vision and information retrieval. As the amount of image data grows explosively, approximate nearest neighbor (ANN) \cite{IndykM98STOC} search becomes a necessary component of CBIR systems, so that images can be retrieved efficiently. Hashing is one of the most popular ANN search approaches. The main idea of hashing based ANN methods is to map images into a similarity-preserved hamming space where the search space can be efficiently pruned.

Deep convolutional neural networks (DCNN) have been successfully applied to the CBIR task \cite{MM14DL4CBIR}, because they can learn better feature representations and be designed as end-to-end models easily. Moreover, a great number of approaches have been proposed to use DCNN for binary code learning \cite{AAAI14SuperHashing, CVPR15DHCompactBinaryCode, CVPR15DeepSemantic, CVPR15CNNH, CVPRW15DLBH, TIP15BitScalableDH, AAAI16DHNetwork, AAAI16DQNetwork, IJCAI16DPSH, IJCAI16DHEncoder, CVPR16DSH, ECCV16BinaryDNN, AAAI17DHJoint, AAAI17TransitiveDH, AAAI17PairwiseDH, IJCAI17DisDH, IJCAI17NolinearDH, IJCAI17LCDH, PAMI17SSDH}. All these papers claim that deep hashing methods outperform the traditional hashing methods such as LSH \cite{GiVLDBIM1999VLDB}.  However, there are serious flaws in the evaluations of these deep hashing papers: 

\begin{enumerate}
	\item In \cite{AAAI14SuperHashing, CVPR15DHCompactBinaryCode, CVPR15DeepSemantic, CVPR15CNNH, CVPRW15DLBH, TIP15BitScalableDH, AAAI16DHNetwork, AAAI16DQNetwork, IJCAI16DPSH, IJCAI16DHEncoder, CVPR16DSH, ECCV16BinaryDNN, AAAI17DHJoint, AAAI17TransitiveDH, AAAI17PairwiseDH, IJCAI17DisDH, IJCAI17NolinearDH, IJCAI17LCDH}, the datasets used are very small and have very limited number of classes. A good performance on these simple datasets cannot guarantee the good performance on real-life CBIR tasks. Some papers \cite{PAMI17SSDH} use the Imagenet dataset \cite{ILSVRC15} which is large. However, a fully supervised setting is used which is also not appropriate.
	
	\item All the ANN search methods (\eg, the hashing methods) sacrifice accuracy for efficiency. Thus, the search time must be reported when we report some accracy measures (mean average precision, precision at $N$ samples, precision at hamming radius $r$, and precision-recall curves). However, most of the existing deep hashing papers \cite{AAAI14SuperHashing, CVPR15DHCompactBinaryCode, CVPR15DeepSemantic, CVPR15CNNH, CVPRW15DLBH, TIP15BitScalableDH, AAAI16DHNetwork, AAAI16DQNetwork, IJCAI16DPSH, IJCAI16DHEncoder, CVPR16DSH, ECCV16BinaryDNN, AAAI17DHJoint, AAAI17TransitiveDH, AAAI17PairwiseDH, IJCAI17DisDH, IJCAI17NolinearDH, IJCAI17LCDH, PAMI17SSDH} failed to do so.
	
	\item The performance of some unsupervised hashing algorithms (\eg, LSH \cite{GiVLDBIM1999VLDB}) can easily be boosted if one uses multiple hash tables, while it is not clear how to use the same trick (multiple tables) for the deep hashing algorithms. This is an important factor should be considered in the evaluation while most of the papers \cite{AAAI14SuperHashing, CVPR15DHCompactBinaryCode, CVPR15DeepSemantic, CVPR15CNNH, CVPRW15DLBH, TIP15BitScalableDH, AAAI16DHNetwork, AAAI16DQNetwork, IJCAI16DPSH, IJCAI16DHEncoder, CVPR16DSH, ECCV16BinaryDNN, AAAI17DHJoint, AAAI17TransitiveDH, AAAI17PairwiseDH, IJCAI17DisDH, IJCAI17NolinearDH, IJCAI17LCDH, PAMI17SSDH} failed to do so.
	
\end{enumerate}

Therefor, the conclusions (the deep hashing methods are superior than the traditional hashing methods) in all these deep hashing papers should be carefully re-examined.

In this paper, we carefully designed the experiments: 1) use the imagenet dataset; 2) use the precision-time curves; 3) use the multiple tables trick for traditional hashing methods. Three state-of-the-art deep hashing methods \cite{CVPRW15DLBH, IJCAI16DPSH, PAMI17SSDH} are compared with the LSH \cite{GiVLDBIM1999VLDB} and IsoH \cite{KongL12NIPS}. Experimental results reveal that the performance of deep hashing methods are inferior to IsoH \cite{KongL12NIPS}, which is a simple unsupervised hashing method.

The goal of this paper is not aiming at proving that some traditional unsupervised hashing methods are better than the deep hashing methods. We only want to show that the claim hold in most of the existing deep hashing papers \cite{AAAI14SuperHashing, CVPR15DHCompactBinaryCode, CVPR15DeepSemantic, CVPR15CNNH, CVPRW15DLBH, TIP15BitScalableDH, AAAI16DHNetwork, AAAI16DQNetwork, IJCAI16DPSH, IJCAI16DHEncoder, CVPR16DSH, ECCV16BinaryDNN, AAAI17DHJoint, AAAI17TransitiveDH, AAAI17PairwiseDH, IJCAI17DisDH, IJCAI17NolinearDH, IJCAI17LCDH, PAMI17SSDH} that deep hashing is superior than traditional hashing should be carefully re-examined.

\section{Search With The Hash Index} \label{sec:search_algorithm}

Most of the deep hashing papers failed to include the search time in the evaluation. One of the reasons may be that it is very natural to think two hashing algorithms spend the same time on retrieving the same number of images with the same code length. However, with careful analysis on search-with-the-hash-index procedure and experiments, we can find that this initial thought is wrong \cite{DCaiRevisit}. \cite{DCaiRevisit} provides a very detailed analysis on how to search with the hash index, we simply restate the main conclusions here. 

A hashing algorithm generating one bit code actually partitions the original image feature space into two parts, the images in one part receive code 1 and the images in the other part receive code 0. When $l$-bits code is used, the hashing algorithm actually partitions the image feature space into $2^l$ parts, which can be named as {\em hash buckets}. Thus, all the images fall into different hash buckets (associated with different binary codes). Ideally, if neighbor images fall into the same bucket or the nearby buckets (measured by the hamming distance of binary codes), the hashing method can efficiently retrieve the nearest neighbors of a query image. 

Suppose the user submit a query and ask for $K$ images in the database, the procedure of search-with-the-hash-index (One can use the parameter $P$ to control the trade-off between efficiency and accuracy, which is the number of images retrieved to the scanning pool) is as follows:  

\begin{enumerate}
	\item Encode the query image into the binary code.
	\item Locate the bucket indexed by the same binary code as the query code (two codes with the hamming distance $0$) in the hash table, and retrieve at most $P$ images in that bucket into the scanning pool. If there are less than $P$ images in that bucket, the hamming distance between the query code and the fetched buckets is increased. This process is carried out recursively until $P$ images are retrieved.
	\item The $P$ images in the scanning pool are sorted according to their distances from the query image and $K$ nearest neighbors are returned. 
\end{enumerate}
The time spent on each step \cite{DCaiRevisit} is as follows:
\begin{enumerate}
	\item \textbf{Coding time}: The time used to convert the query image to the binary code.
	\item \textbf{Locating time}: The time used to locate the buckets in the hash table and retrieve $P$ candidate images.
	\item \textbf{Scanning time}: The time used to scan images in the scanning pool and return $K$ results.
\end{enumerate}

The coding time and scanning time (proportional to the images in the scanning pool) can be easily analyzed. However, most of the existing deep hashing papers ignored the locating time which becomes the dominate part when one aims at high accuracy \cite{DCaiRevisit}. 

Given a binary code $q$, locating the hashing bucket corresponding to $q$ costs $O(1)$ time (by using std::vector or std::unordered\_map). If we only need to examine a small number of hash buckets, the locating time can be neglected. This happened if all the neighbor images fall into the same bucket or the nearby buckets (measured by the hamming distance of binary codes) ideally. However, there is no guarantee that all the neighbor images will fall into the nearby buckets. To ensure a high precision, one needs to examine many hashing buckets (\ie, enlarge the search radius in hamming space). 

Given a  $l$-bits binary code $q$, considering those hashing buckets whose hamming distance to $q$ is small or equal to $r$. It is easy to show the number of these buckets is $\sum_{i=0}^r\binom{l}{i}$, which increases almost exponentially with respect to $i$.
Thus, the locating time can not be ignored if we aim at achieving a high precision (\ie, we need to examine many hash buckets).

Even with the same code length and with the same parameter $P$, two different hashing algorithms can lead to significant different locating time. This main due to the different distributions of images over the hamming space. Indeed the locating time is highly related to the quality of the binary codes, and it should never be ignored when evaluating hashing methods for CBIR.

\section{Deep Hashing VS. Traditional Hashing} 

\begin{figure}[t]
	\centering
	\subfigure[AlexNet]{
		\includegraphics[width=\SmallSingleFigureWidth]{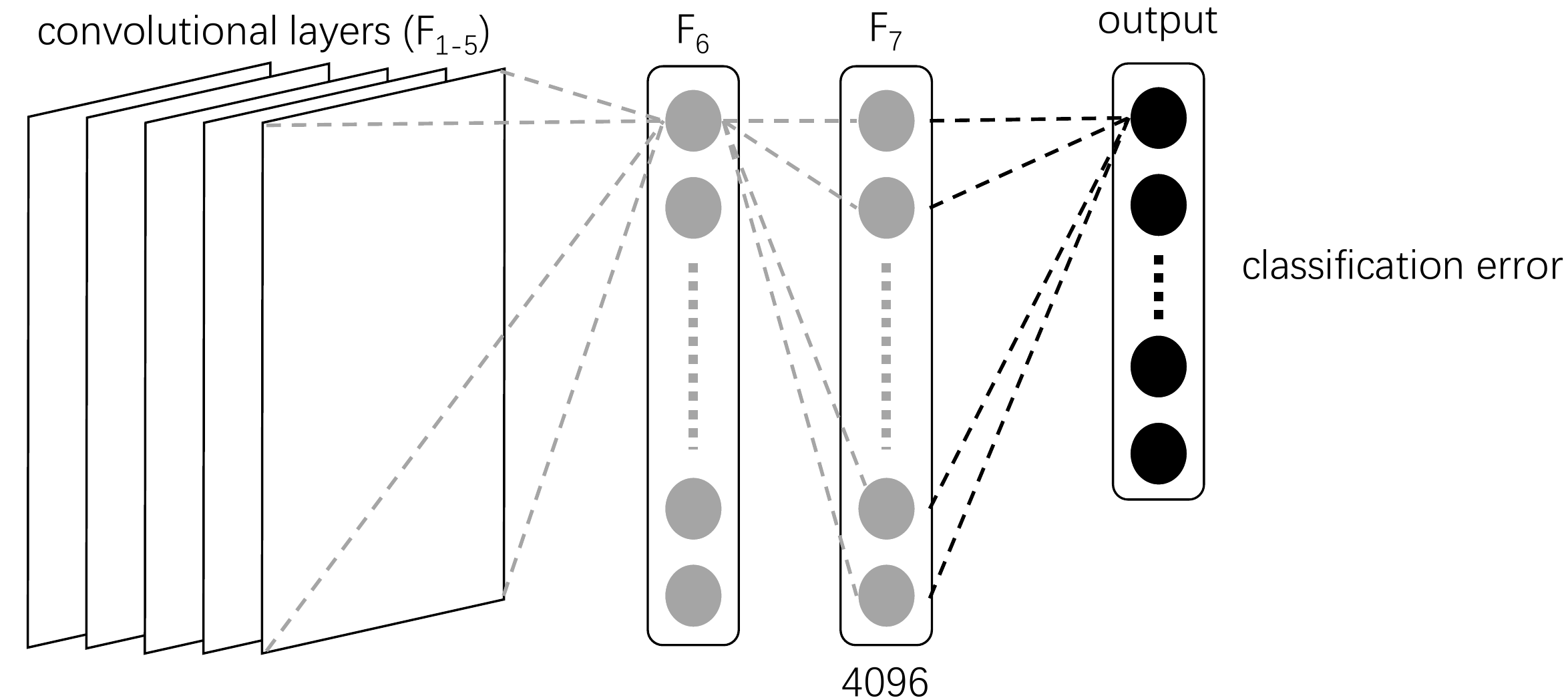}}
	\subfigure[Deep Hashing Net (SSDH \cite{PAMI17SSDH})]{
		\includegraphics[width=\SingleFigureWidth]{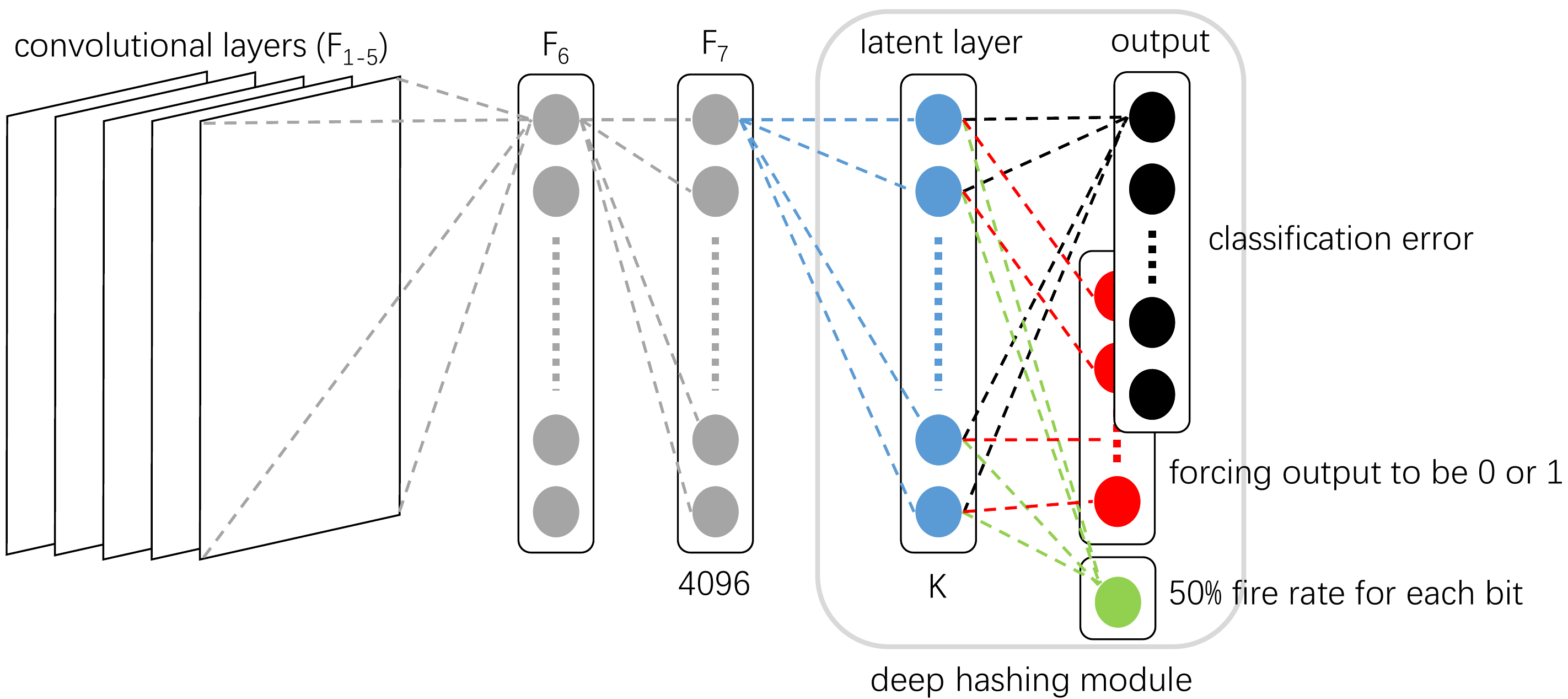}}
	\caption{An illustration of the deep hashing model. A deep hashing model (b) can be modified from any existing deep model (a) by replacing the output layer to various deep hashing modules.}
	\label{fig:deephashingnet}
\end{figure}

Inspired by the revolutionary success of DCNN on the computer vision tasks, researchers proposed to use DCNN for binary code learning, so called deep hashing methods \cite{AAAI14SuperHashing, CVPR15DHCompactBinaryCode, CVPR15DeepSemantic, CVPR15CNNH, CVPRW15DLBH, TIP15BitScalableDH, AAAI16DHNetwork, AAAI16DQNetwork, IJCAI16DPSH, IJCAI16DHEncoder, CVPR16DSH, ECCV16BinaryDNN, AAAI17DHJoint, AAAI17TransitiveDH, AAAI17PairwiseDH, IJCAI17DisDH, IJCAI17NolinearDH, IJCAI17LCDH, PAMI17SSDH}. The deep models for deep hashing learning are not complicated. Actually, these deep hashing models can be modified from any conventional deep model (for classification tasks, \eg, AlexNet \cite{alexnet}, VGG-16 \cite{VGG-16}, GoogLeNet \cite{GoogLeNet} and ResNet \cite{resnet}) simply by replacing the output layer to various {\em deep hashing modules} (for the purpose of various loss functions, ensuring the binary outputs, \etc). Figure \ref{fig:deephashingnet} gives an example of converting a conventional deep model (we use AlexNet as an example) to a deep hashing model (we use SSDH \cite{PAMI17SSDH} as an example). The gray box part in Figure \ref{fig:deephashingnet} (b) shows the deep hashing module used in SSDH \cite{PAMI17SSDH}. 

For the representation purpose, both the $F_7$ layer in the AlexNet (Figure \ref{fig:deephashingnet} (a)) and the $F_7$ layer in the Deep Hashing Net (Figure \ref{fig:deephashingnet} (b)) provide the 4096-dimensional real vector representation of the input image. With this real vector representation, the traditional hashing algorithms (\eg, LSH \cite{GiVLDBIM1999VLDB}) can then be applied for binary code learning. One of the main motivations of deep hashing methods is the joint learning of representation and binary code could leads to better binary codes \cite{AAAI14SuperHashing, CVPR15DHCompactBinaryCode, CVPR15DeepSemantic, CVPR15CNNH, CVPRW15DLBH, TIP15BitScalableDH, AAAI16DHNetwork, AAAI16DQNetwork, IJCAI16DPSH, IJCAI16DHEncoder, CVPR16DSH, ECCV16BinaryDNN, AAAI17DHJoint, AAAI17TransitiveDH, AAAI17PairwiseDH, IJCAI17DisDH, IJCAI17NolinearDH, IJCAI17LCDH, PAMI17SSDH}.

However, this motivation should be carefully examined. Since the binary codes are learned from the real vector representation, the goodness of the real vectors are one of the keys determining the performance of the binary codes. Thus, it is unfair to compare deep hashing methods with traditional hashing algorithms with hand-craft features as the inputs \cite{AAAI14SuperHashing, AAAI16DHNetwork, AAAI16DQNetwork, IJCAI17NolinearDH}. Moreover, with the deep hashing module, it has high possibility that the 4096-dimensional real features of the $F_7$ layer in the AlexNet (Figure \ref{fig:deephashingnet} (a)) will be different to the 4096-dimensional real features of the $F_7$ layer in the Deep Hashing Net (Figure \ref{fig:deephashingnet} (b)), although both two deep networks share the same structure in the formal part. Thus, it is also unfair to compare the deep hashing codes learned from the Figure \ref{fig:deephashingnet} (b) and the traditional hashing algorithms with the input from the $F_7$ layer in the Figure \ref{fig:deephashingnet} (a). A fare comparison should let the traditional hashing algorithms take the inputs from the $F_7$ layer in the Figure \ref{fig:deephashingnet} (b).

Another important difference between deep hashing and traditional hashing is that many traditional unsupervised hashing algorithms can use the so called  multiple hash tables trick \cite{DCaiRevisit} with almost no additional computational burdon. Take LSH \cite{GiVLDBIM1999VLDB} as an example. Since LSH is essentially based on random projection, two hash tables generate by LSH naturally will be different (\ie, a query point will have different neighbor vectors in nearby hash buckets). To locate $P$ points, if we only have one hash table, we have to increase the hamming radius $r$ if the points in all the buckets within the hamming distance $r$ are not enough. This will increase the locating time a lot. If we have multiple hash tables, instead of increasing $r$, we can scan the buckets within the hamming radius $r$ in all the tables, which gives us a larger chance to locate enough data points. There are plenty experimental results in \cite{DCaiRevisit} show the superior performance by using the multiple tables trick.

To use the multiple tables trick, a hashing algorithm must generate different hash tables for the same dataset. Some hashing algorithms (\eg, LSH \cite{GiVLDBIM1999VLDB} and IsoH \cite{KongL12NIPS}) have randomness in nature and naturally can use the multiple tables trick. Meanwhile, there is no need to change the inputs (4096-dimensional real vectors) for these traditional hashing algorithms, \ie, there is no need to train multiple deep models. During the training stage, since the training process for many traditional hashing algorithms (\eg, LSH \cite{GiVLDBIM1999VLDB} and IsoH \cite{KongL12NIPS}) are very efficient, there is almost no additional computational burden on training. More importantly, during the search stage, using LSH or IsoH to convert a real vector to a binary vector is extremely fast, the coding time will be almost the same for single table or multiple tables (see next section for a detailed analysis). 

For a deep hashing method, since the deep model usually converges at a local optimum, it is possible that two times of training generate different deep hashing codes. However, this means we need to train the deep model several times (depend on how many hash tables we want to use). This process will introduce significant amount of computational burden on the training stage for a large scale data. More importantly, feedforward the query image through multiple deep networks increase the coding time significantly. This makes the multiple trick cannot be used for all the deep hashing methods.

\section{Experiments} 

In the remaining part of the paper, we will perform extensive experiments to support our finding. We begin with the description on the datasets we used in the experiments. 

\subsection{Datasets} 
Three datasets are used in this paper. Two of them are small and one is large.

\begin{itemize}
	\item CIFAR-10 which contains 600,000 images belonging to 10 categories.
	\item MNIST which contains 700,000 images belonging to 10 categories.
	\item Imagenet which contains more than 1.2M images belonging to 1,000 categories.
\end{itemize}

\subsection{Compared Methods} 

Three deep hashing methods are compared in the experiments, they are: 

\begin{itemize}
	\item {\bf DLBH} in the paper of {\em Deep Learning of Binary Hash codes for fast image retrieval} \cite{CVPRW15DLBH}.
	\item {\bf DPSH} in the paper of {\em feature learning based Deep Supervised Hashing with Pairwise labels} \cite{IJCAI16DPSH}.
	\item {\bf SSDH} in the paper of {\em Supervised learning of Semantics-preserving Hash via Deep convolutional neural networks} \cite{PAMI17SSDH}.
\end{itemize}
The main reason to pick these thee methods is all of them provide publicly available codes\footnote{https://github.com/kevinlin311tw/caffe-cvprw15} \footnote{http://cs.nju.edu.cn/lwj/code/DPSH.zip} \footnote{https://github.com/kevinlin311tw/Caffe-DeepBinaryCode}.

As we have discussed in the Section 3, a deep hashing model can be modified from any conventional deep model for the classification task.
To make fair comparisons, all the three deep hashing models are adapted from the ResNet \cite{resnet} architecture. 
Our implementation is based on the torch ResNet implementation\footnote{https://github.com/facebook/fb.resnet.torch/tree/master}. The deep hashing modules for  three deep hashing methods are strictly follow the original implementations in the publicly available codes. 

We use the ResNet-34 network for the two small datasets and the ResNet-50 network for the Imagenet dataset. All the images are resize to 224*224 in order to fit the ResNet input size.

Two traditional unsupervised hashing methods are compared in the experiments, they are: 
\begin{itemize}
	\item {\bf LSH} is a short name for Locality Sensitive Hashing \cite{GiVLDBIM1999VLDB}. LSH is based on random projection and is frequently used as a baseline method in various hashing papers.
	\item {\bf IsoH} is a short name for Isotropic Hashing \cite{KongL12NIPS}. IsoH learns the
	projection functions with isotropic variances for PCA projected	data. The main motivation of IsoH is that PCA directions with different variance should not be equally treated (one bit for one direction). The performance of IsoH is quite good according to the comparative study in \cite{DCaiRevisit}.
\end{itemize}
Both algorithms can be downloaded at github\footnote{https://github.com/dengcai78/MatlabFunc/tree/master/ANNS/Hashing}.

One reason to choose these two hashing algorithms is the efficiency of these two algorithms in the test stage. Both algorithms only need a matrix multiplication to convert the input vector to the required binary code. 
Thus, the LSH and IsoH can also be modeled using the deep network in Figure \ref{fig:deephashingnet} (a). After the network learning, one can simply replace the weights connecting the layer $F_7$ and the output layer by the transformation matrix learned in LSH (or IsoH).

We report the performance of LSH and IsoH using single table and 16 tables. For deep hashing methods, we only report the performance using single table. The reason is that it is too time consuming to learn multiple deep hashing tables. Take the Imagenet dataset as an example. Train a deep hashing model modified from the ResNet-50 network with only 10\% training data on a 4 GTX1080 GPU machine costs more than 20 hours. In contract, train a LSH table from learned deep features only needs 2.09s and train a IsoH table only needs 35.16s on an i7-5930K CPU. Moreover, using multiple tables for deep hashing methods needs to keep multiple deep models. The coding time of using multiple deep hashing tables will be extremely longer (every query image has to feedforward through multiple deep models), see the next subsection for a detailed analysis.

To make fair comparisons, we use the features extracted from the fully-connected layer immediately before the layer that generates binary codes of DLBH as the inputs to LSH and IsoH. The dimensionality of the feature is 512 for the two small datasets and 2048 for the Imagenet dataset.

\begin{figure*}[t]
	\centering
	\subfigure[Precision@10]{
		\includegraphics[width=\DoubleFigureWidth]{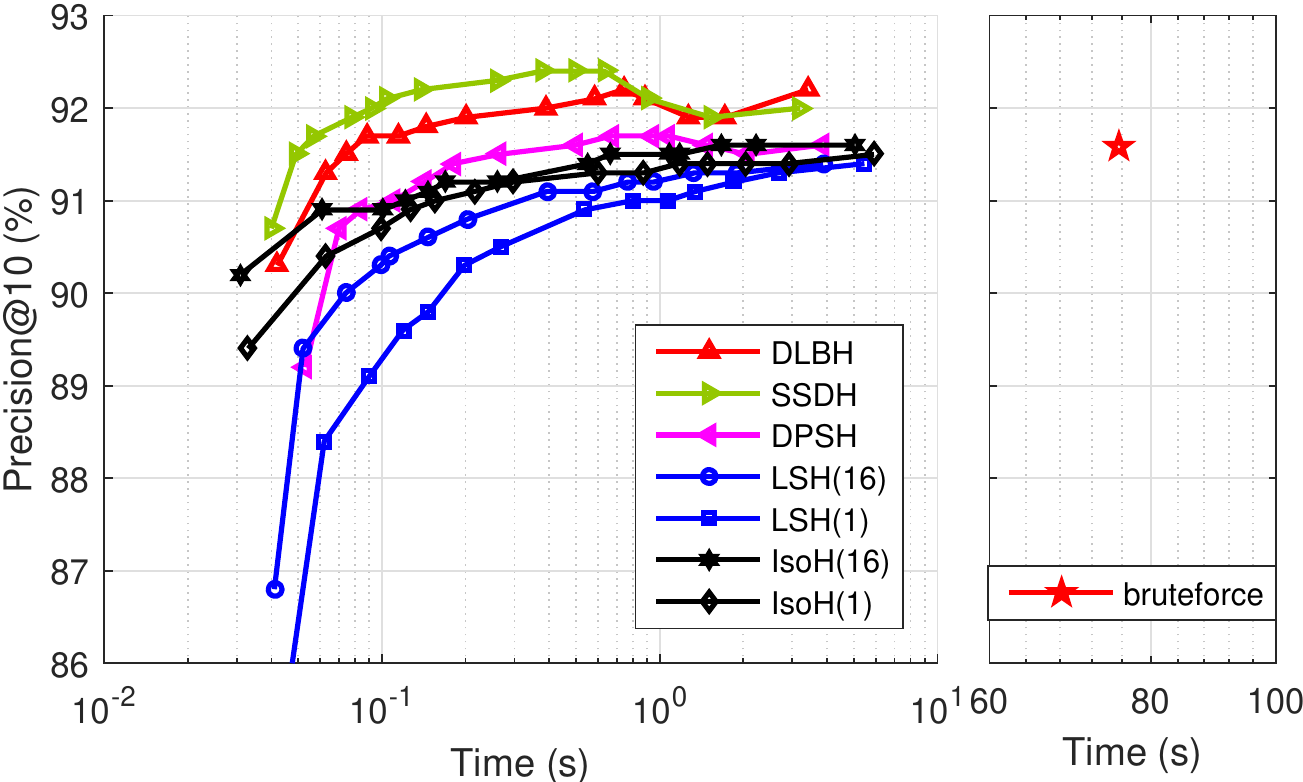}}
	\subfigure[Precision@100]{
		\includegraphics[width=\DoubleFigureWidth]{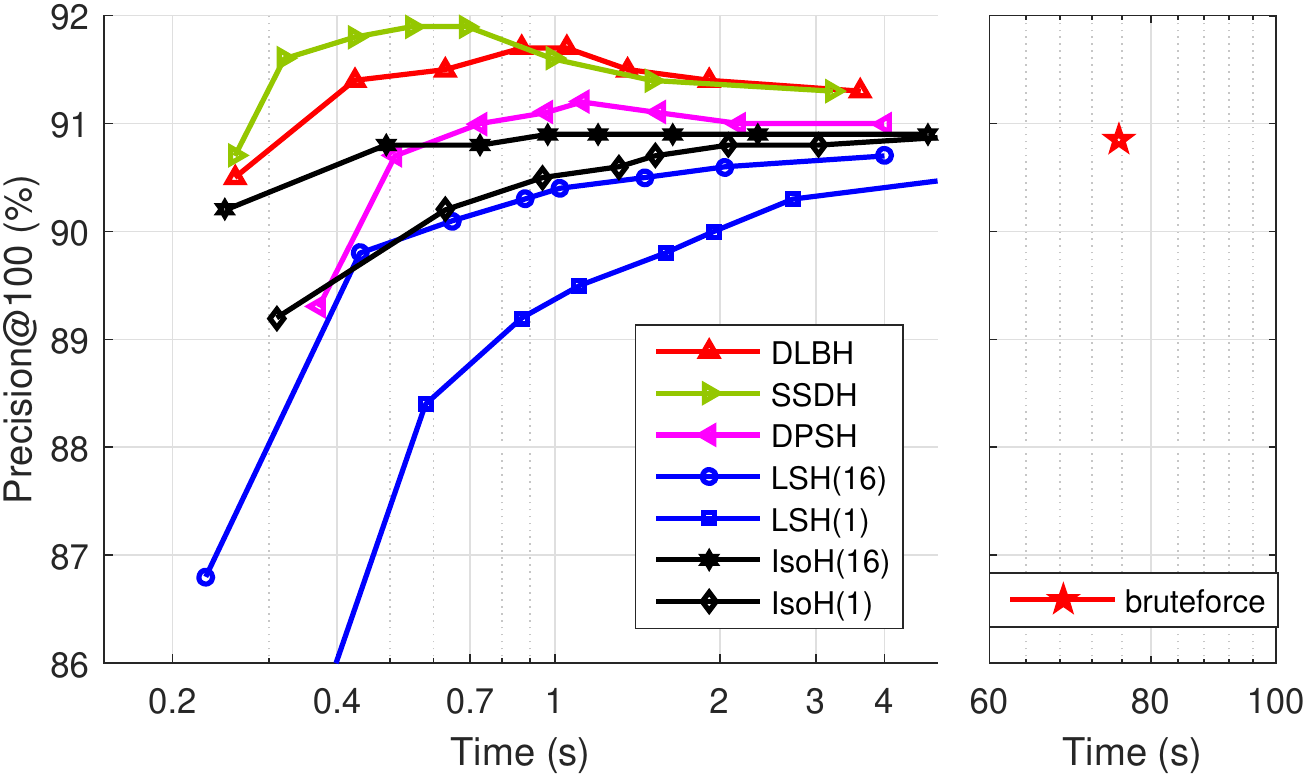}}
	\caption{Time-Precision curves of various hashing algorithms on CIFAR-10.}
	\label{fig:res_CIFAR10}
\end{figure*}

\subsection{Evaluation Criteria} 

Given a query image, the algorithm is expected to return $K$ images. Then we need to examine how many images in this returned set are relevant to the query image (share the same label). Suppose the returned set of relevant images given a query is $R$, the $precision@K$ can be defined \cite{Makhoul2000Performance} as
\begin{equation}
precision@K = \frac{|R|}{K}.
\end{equation}
We fixed $K=10$ and $K=100$ throughout our experiments.

As we have discussed in the section 2, the search time should be reported as well as the precision. Thus, we use the {\em precision-time curve} for evaluation. Ideally, the search time should include all the three parts: coding time, locating time and scanning time. 

The coding time is the time used to convert the query image to the binary code. This can be divided as two parts. The first part is converting the image to the real vector feature (from the input to $F_7$ layer in the Figure \ref{fig:deephashingnet} (b)), which is the same for all the hashing methods (both deep hashing methods and traditional hashing methods). The second part is converting the real vector feature to binary code (deep hashing methods use the deep hashing module in Figure \ref{fig:deephashingnet} (b) while the traditional hashing use a matrix multiplication). Since the time spent on the first part is the dominating part in the coding time, the coding time for deep hashing methods and LSH (or, IsoH) are almost the same. 

\begin{figure*}[t]
	\centering
	\subfigure[Precision@10]{
		\includegraphics[width=\DoubleFigureWidth]{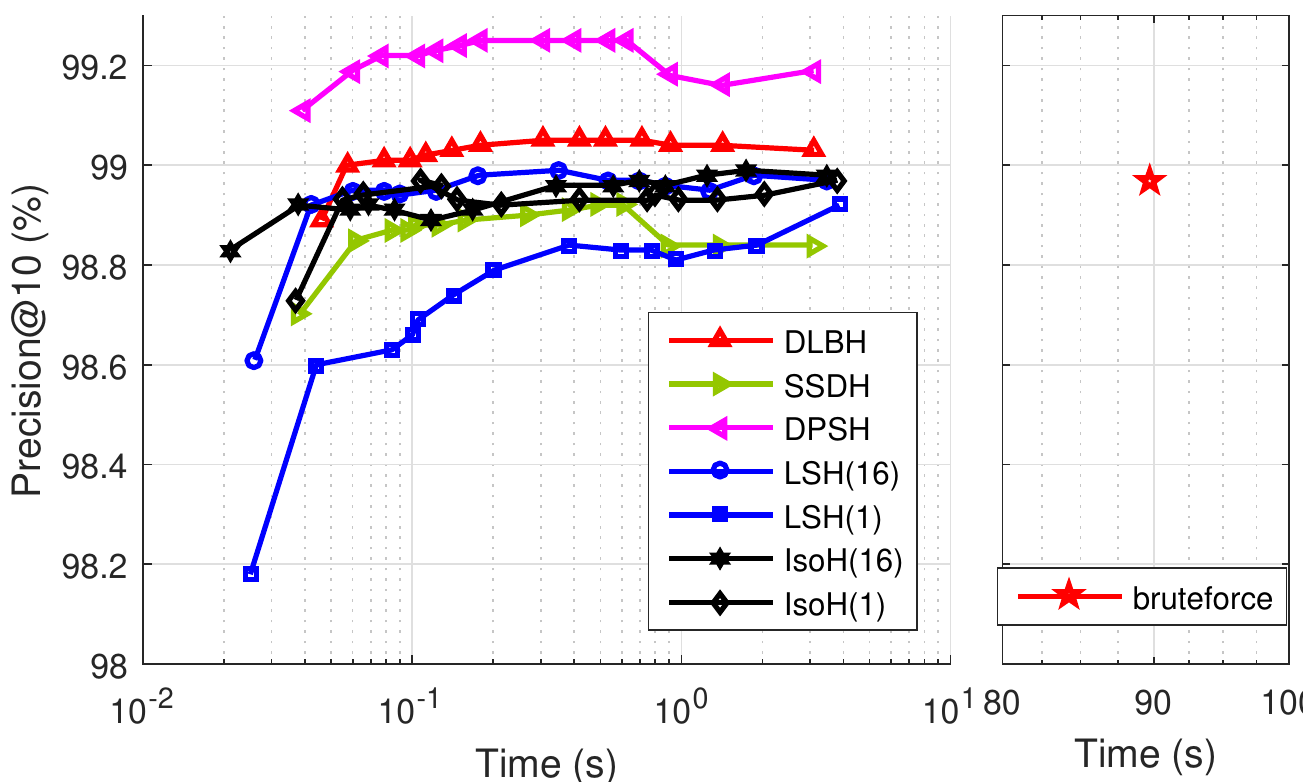}}
	\subfigure[Precision@100]{
		\includegraphics[width=\DoubleFigureWidth]{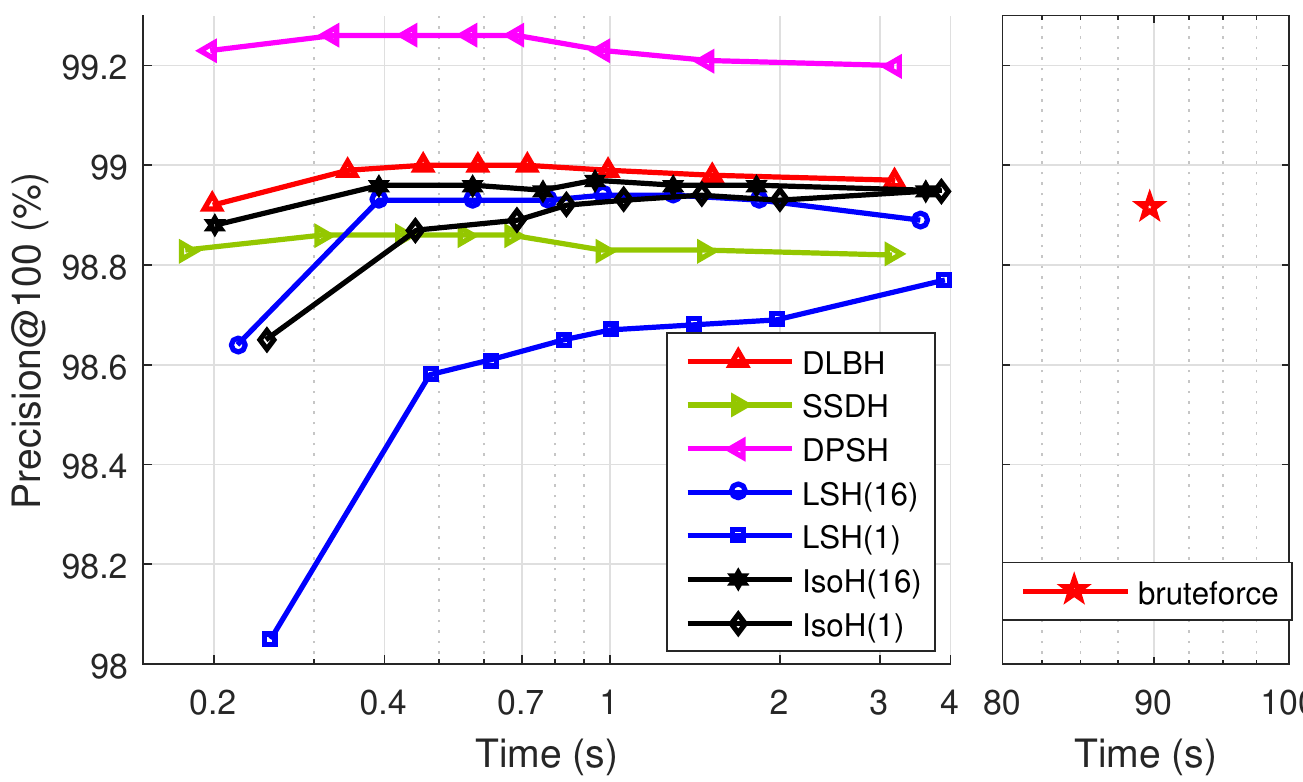}}
	\caption{Time-Precision curves of various hashing algorithms on MNIST.}
	\label{fig:res_MNIST}
\end{figure*}

Take the Imagenet as an example, based on the study on github\footnote{https://github.com/jcjohnson/cnn-benchmarks}, if the ResNet-50 network is used, the first part of the coding time is around 50ms per image on GTX1080. If we have 25,000 query images, the total time for the first part is around 125s. While the total processing time (the second part of the coding time) for LSH (or IsoH) for 25,000 query images is around 0.037s on an i7-5930K CPU. Even we need 16 hash tables for LSH (or IsoH), the time on the second part can still be neglected compared to the time on the first part.

Thus, we can safely conclude that the coding time for deep hashing methods and LSH (or IsoH) are almost the same, even if LSH (or IsoH) uses 16 hash tables. 
Moreover, our primary goal is evaluating the quality of the binary codes generated by different hashing methods. This will be mainly reflected by the locating and scanning time.

Thus, the time in {\em precision-time curve} reported in our experiments will only include the locating time and scanning time. After obtaining the codes generated by different hashing methods, we use the open source c++ {\em search with the hash index} code\footnote{https://github.com/fc731097343/efanna/tree/master/samples\_hashing} on the same i7-5930K CPU for fair evaluation (by tuning the parameter $P$, we can get the curves for different hashing methods).

\begin{table} 
	\captionsetup{aboveskip=-10pt}
	\caption{The data split on CIFAR-10 and MNIST} \label{table:smallDatasetSplits}
	\begin{center}
		\begin{tabular}{|l|c|c|c|c|}
			\hline
			Dataset & \#Train & \#Val & \#Base & \#Query \\
			\hline\hline
			CIFAR-10 & 5000 & 5000 & 50000 & 5000 \\
			MNIST & 6000 & 5000 & 60000 & 5000 \\
			\hline
		\end{tabular}
	\end{center}
	The images in Val and Query sets are from the test split provided by the original dataset. The images in Base set are from the train split provided by the original dataset. The images in Train set are 10\% randomly selected from the Base set. Thus, the images in Val, Query and Base sets are different from each other.
\end{table}

\subsection{Experiments on Small Datasets}

In this section, we report the experimental results on two small datasets, \ie, CIFAR-10 and MNIST. 

Based on the datasets' original train/test splits, we build our base/train/validation/query splits. We use the train/validation sets to train the deep hashing networks, and use the base/query sets to simulate the image search. On both two datasets, we use the original train set as our base set, and randomly select 10\% images from each category in the base set to form our train set. We then split each category of the original test set evenly and randomly into our validation and query sets, so that our models cannot see the query images when training to learn the binary codes. More details of each set's size are recorded in the Table \ref{table:smallDatasetSplits}.

Figure \ref{fig:res_CIFAR10} and \ref{fig:res_MNIST} show the performances of various hashing methods on CIFAR-10 and MINST respectively. We use 24-bit code for all the hashing methods (Please see the supplementary file for detailed discussion on the selection of the code length).

We can clearly see the performance boost by using 16 tables than single table on LSH and IsoH at all the cases. However, even 16 tables are used, the best deep hashing method still achieve significant better results on both datasets. This result is consistent with the previous deep hashing papers.

However, CIFAR-10 and MNIST are too small and the category number is only 10. The results on these two small datasets are not enough to convince people that best method so far will perform the best on a real large scale complicated CBIR system. 

\begin{figure*}[t]
	\centering
	\subfigure[Precision@10]{
		\includegraphics[width=\DoubleFigureWidth]{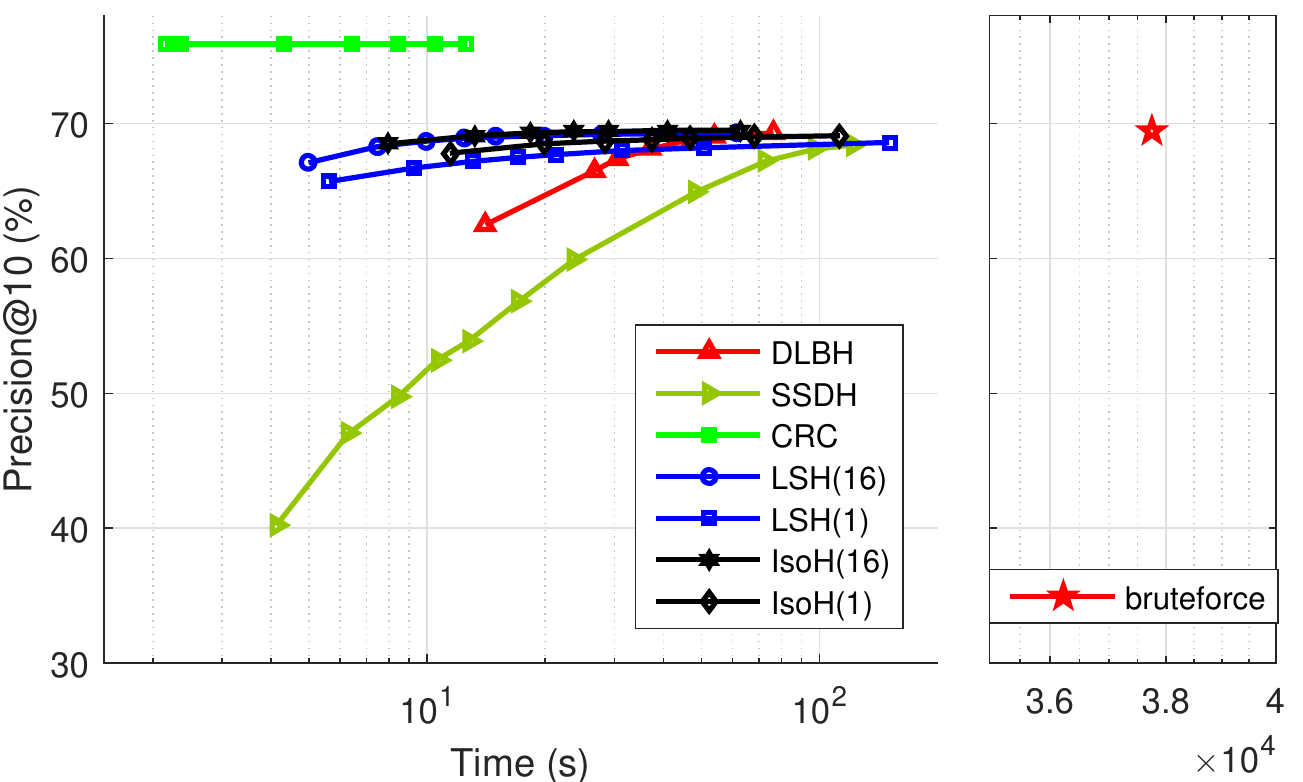}}
	\subfigure[Precision@100]{
		\includegraphics[width=\DoubleFigureWidth]{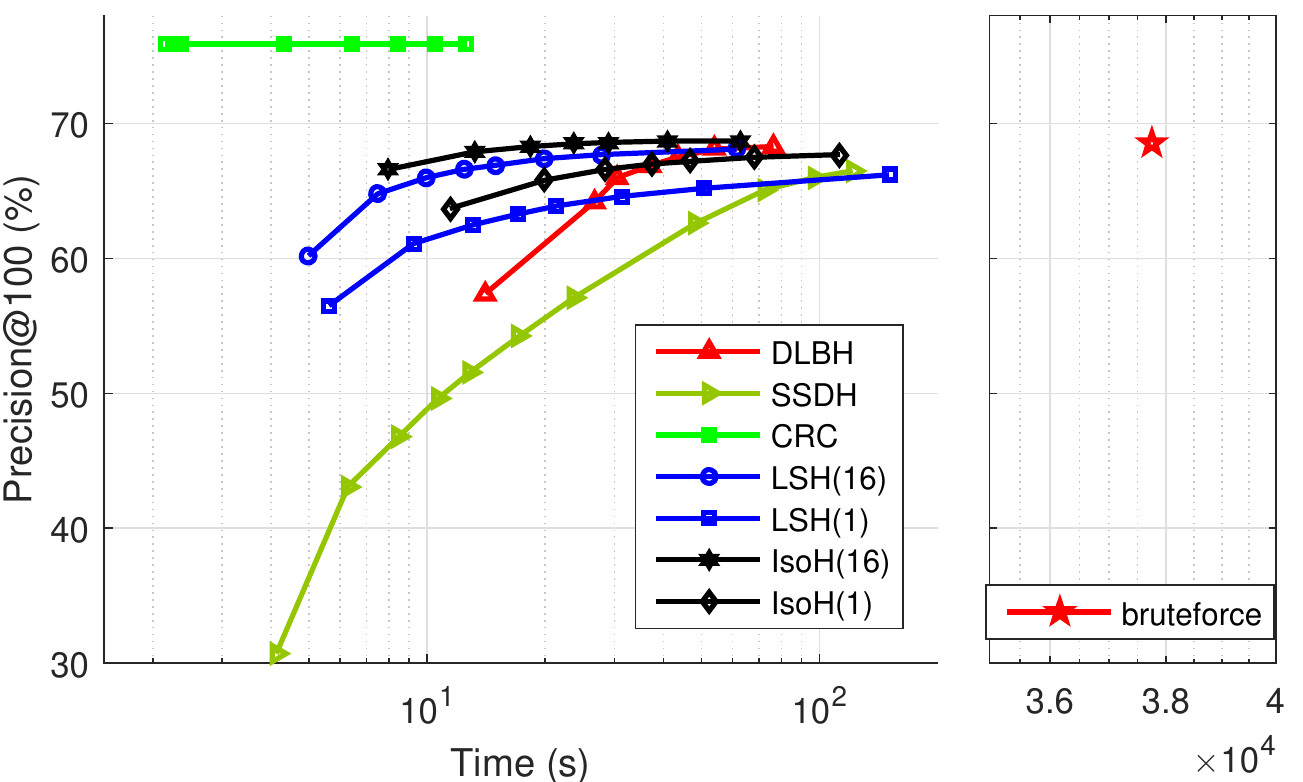}}
	\caption{Time-Precision curves of various hashing algorithms on fully supervised Imagenet.}
	\label{fig:res_ILSVRC15-F}
\end{figure*}

\begin{figure*}[t]
	\centering
	\subfigure[Precision@10]{
		\includegraphics[width=\DoubleFigureWidth]{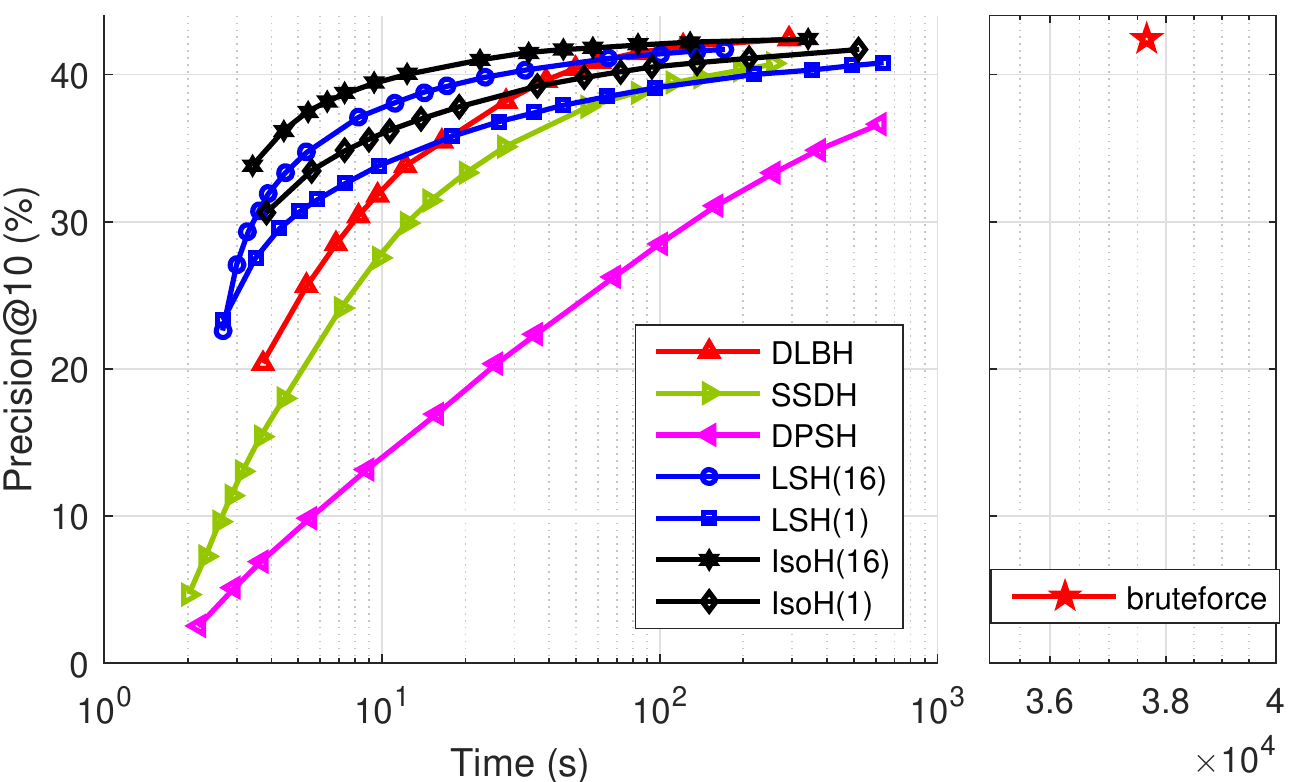}}
	\subfigure[Precision@100]{
		\includegraphics[width=\DoubleFigureWidth]{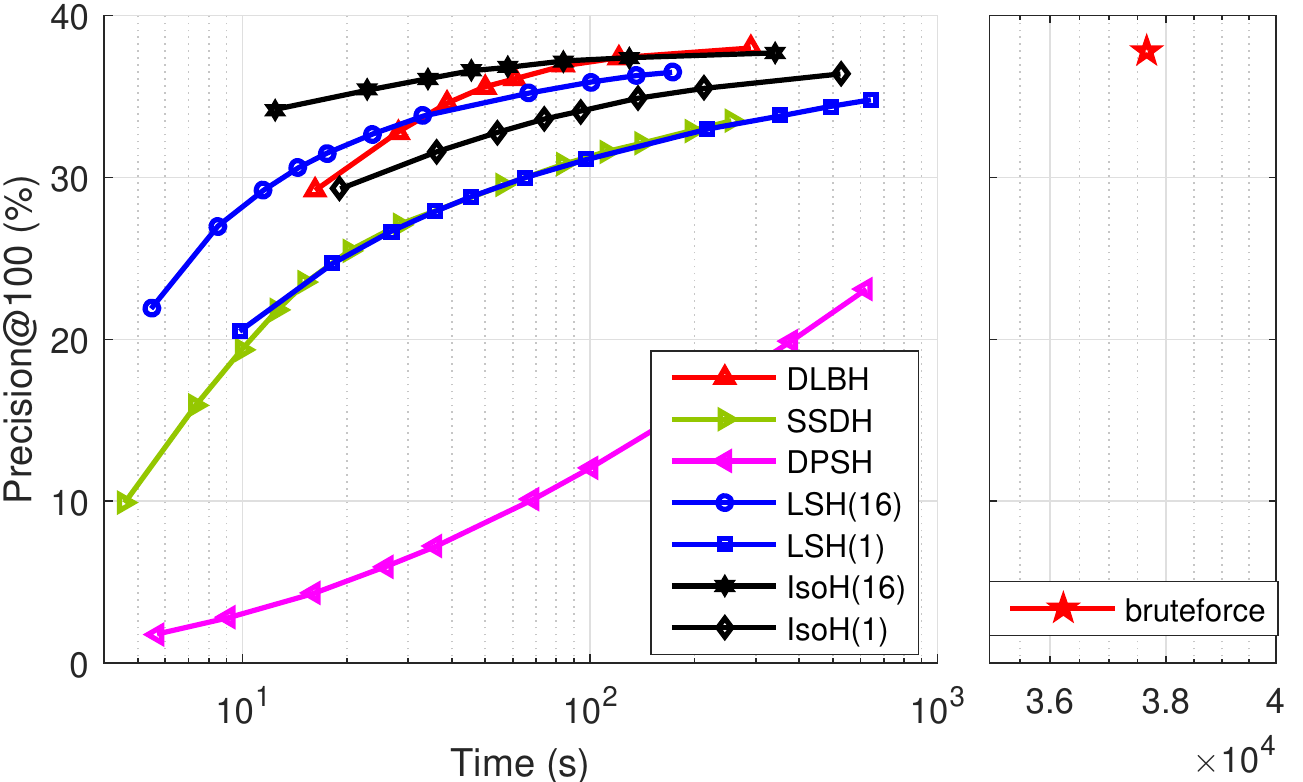}}
	\caption{Time-Precision curves of various hashing algorithms on partially supervised Imagenet.}
	\label{fig:res_ILSVRC15-P}
\end{figure*}

\subsection{Experiments on Imagenet} 

We need to use a more complicated and larger dataset. The Imagenet dataset \cite{ILSVRC15} is a good choice. It contains more than 1.2M images belonging to 1,000 categories.

\subsubsection{Fully supervised setting} 

Actually, the Imagenet dataset has been used in the SSDH paper \cite{PAMI17SSDH}. However, the labels of all the images in the base set are used for learning the binary codes.  In real-life CBIR scenarios, the image database is very large, and it will be too expensive and time-consuming to manually label all the images in the base set. Moreover, with all the label information of the base set available, one can use a very simple but effective coding scheme which makes all other hashing algorithms meaningless.

\begin{table} 
	\captionsetup{aboveskip=-10pt}
	\caption{The data split on fully supervised Imagenet} \label{table:full_imagenet_dataset_setting}
	\begin{center}
		\begin{tabular}{|l|c|c|c|c|}
			\hline
			Dataset & \#Train & \#Val & \#Base & \# Query \\
			\hline\hline
			Imagenet & 1281167 & 25000 & 1281167 & 25000 \\
			\hline
		\end{tabular}
	\end{center}
	The images in Val and Query sets are from the validation split provided by the original dataset. The images in Train and Base set are the same, come from the train split provided by the original dataset. Thus, the images in Val, Query and Base sets are different from each other.
\end{table}

{\bf Classification random coding}: 
With the label information of the image in the entire base set available, we can design
a simple binary encoding scheme with the help of a well trained
classifier: suppose the code length is $l$, each class in the dataset is uniquely and randomly mapped to an integer ranging from $0$ to $2^{l} - 1$, and then the integer is converted to its binary representation of length $l$. Thus, each class maps to a unique binary code. The images in the base set can be simply encoded using the labels and the query images can be encoded using the predicted labels obtained from the classifier. We denote this coding scheme as {\em classification random coding} (CRC).

With this coding scheme, there are only $c$ (the number of classes) non empty buckets, and each bucket contains all the base images belonging to the corresponding class. In the search stage, if the required number of returns is smaller then the number of images in one class (which usually is the case), one only need to locate one hash bucket (with the hamming distance 0 to the binary code of the query image). This means the locating time of CRC can almost be ignored. Moreover, if the classifier correctly predicts the class label of the query image, all the returned results will be relevant and the precision is 100\%. If the classifier misclassified the query image, the precision then is 0. Then the average search precision will be the same as the accuracy of the classifier. If the classifier is good, by using CRC, one can achieve a very good search precision with a very short amount of time.

Figure \ref{fig:res_ILSVRC15-F} shows the performance of all the compared hashing methods, including CRC, on the fully supervised Imagenet. We have 25,000 query images and please see the Table \ref{table:full_imagenet_dataset_setting} for more details of each set's size. All the hashing methods use 32-bit code. The DPSH method \cite{IJCAI16DPSH} is missed in the figure. We tried all the combination of the parameters but the DPSH model failed to converge. This probably due to there are too many of training images. 

Again, 16 tables LSH (or, IsoH) is better than its single table version. IsoH is slightly better than LSH considering Precision@10. If we consider Precision@100, the advantage of IsoH over LSH becomes clear which is consistent with the finding in \cite{DCaiRevisit}.

The improvement of 16 tables IsoH over the other two deep hashing methods is significant which makes the conclusion in most of the deep hashing papers doubtful.

CRC achieves the best performance. It is interesting to find that the precisions of all the other hashing methods converge to the precision of the bruteforce search. While the precision of CRC is significantly better than that of the bruteforce search. The reason might be the performance of the bruteforce search is somehow like the nearest neighbor classifier on deep features while the precision of CRC equals the accuracy of a linear classifier on deep features.

\begin{table} 
	\captionsetup{aboveskip=-10pt}
	\caption{The data split on partially supervised Imagenet} \label{table:10p_imagenet_dataset_setting}
	\begin{center}
		\begin{tabular}{|l|c|c|c|c|}
			\hline
			Dataset & \#Train & \#Val & \#Base & \#Query  \\
			\hline\hline
			Imagenet & 128116 & 25000 & 1281167 & 25000 \\
			\hline
		\end{tabular}
	\end{center}
	The images in Val and Query sets are from the validation split provided by the original dataset. The images in Base set are from the train split provided by the original dataset. The images in Train set are 10\% randomly selected from the Base set. Thus, the images in Val, Query and Base sets are different from each other.
\end{table}

\begin{figure*}[h]
	\centering
	\subfigure[]{\includegraphics[width=145pt]{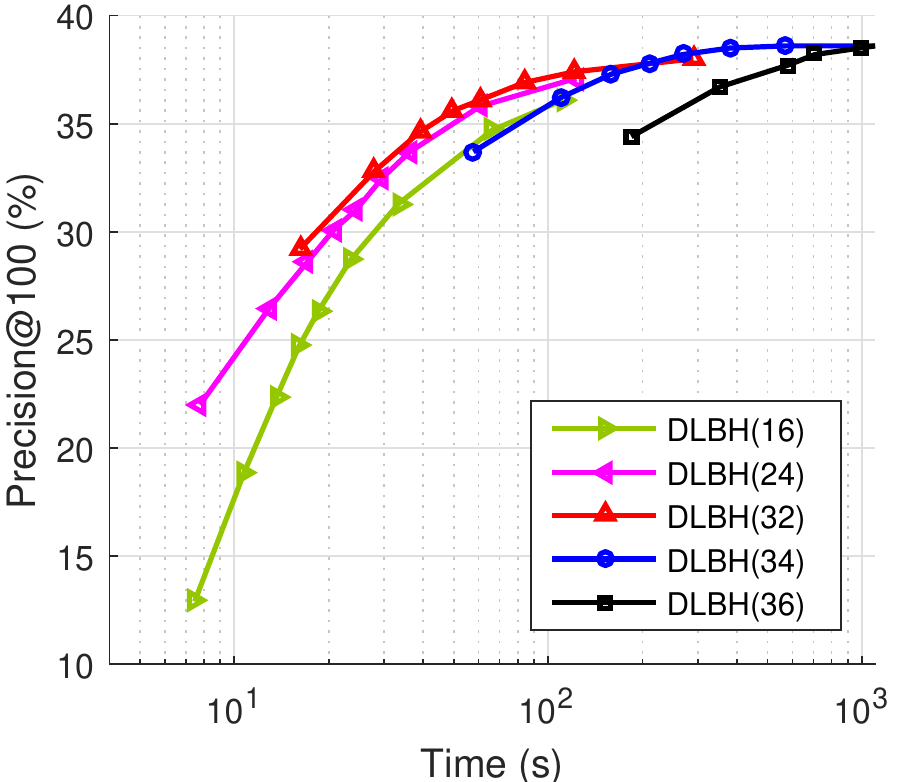}\hspace{20pt}}
	\subfigure[]{\includegraphics[width=145pt]{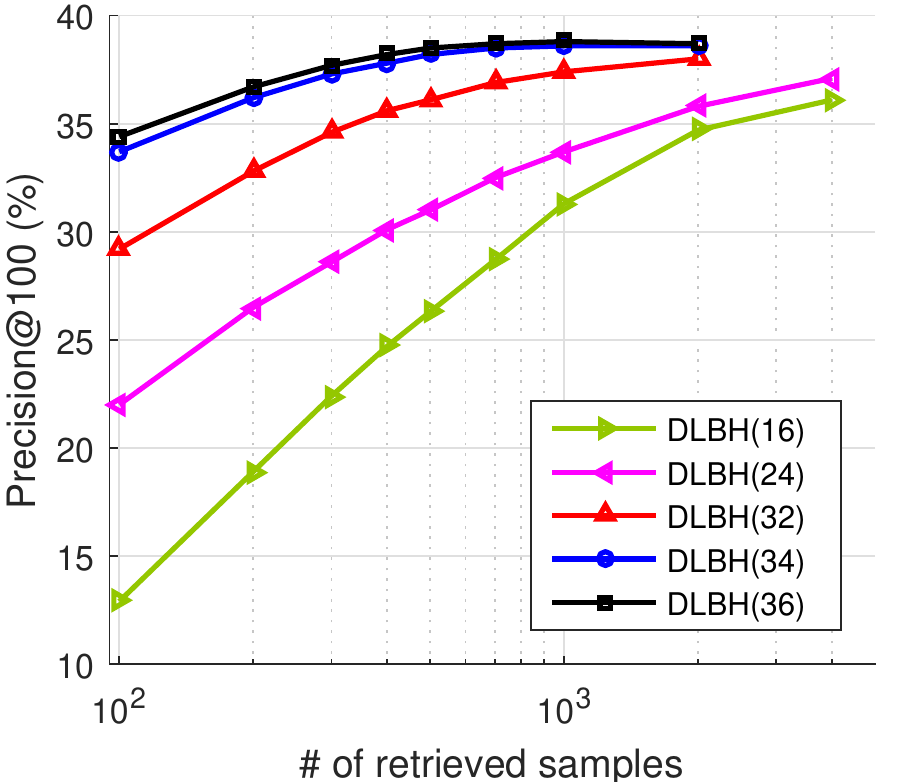}}
	\caption{Precision@100 of DLBH with different code length on partially supervised Imagenet}
	\label{fig:code_length}
\end{figure*}

\begin{figure*}[t]
	\centering
	\subfigure[16 bit]{\includegraphics[width=\FifthFigureWidth]{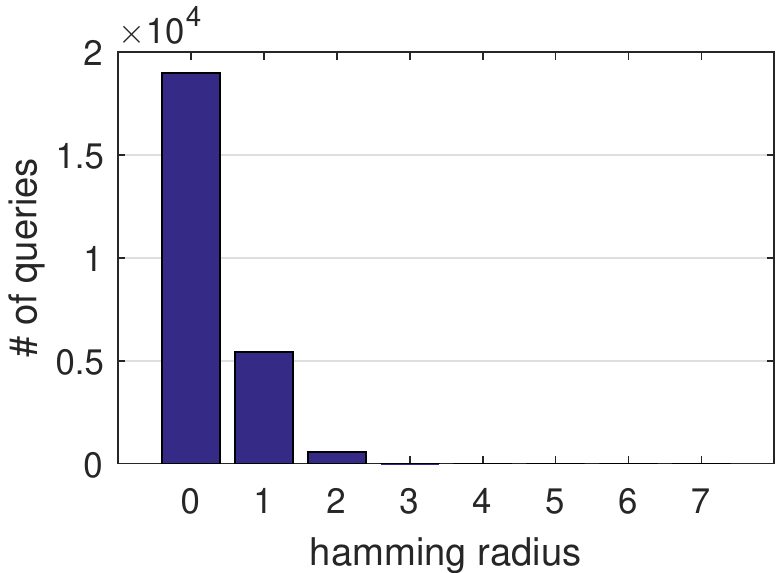}}
	\subfigure[24 bit]{\includegraphics[width=\FifthFigureWidth]{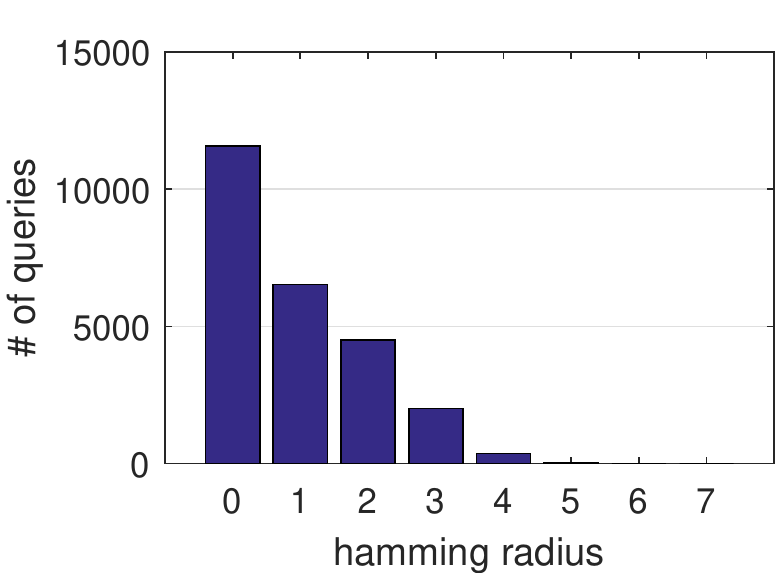}}
	\subfigure[32 bit]{\includegraphics[width=\FifthFigureWidth]{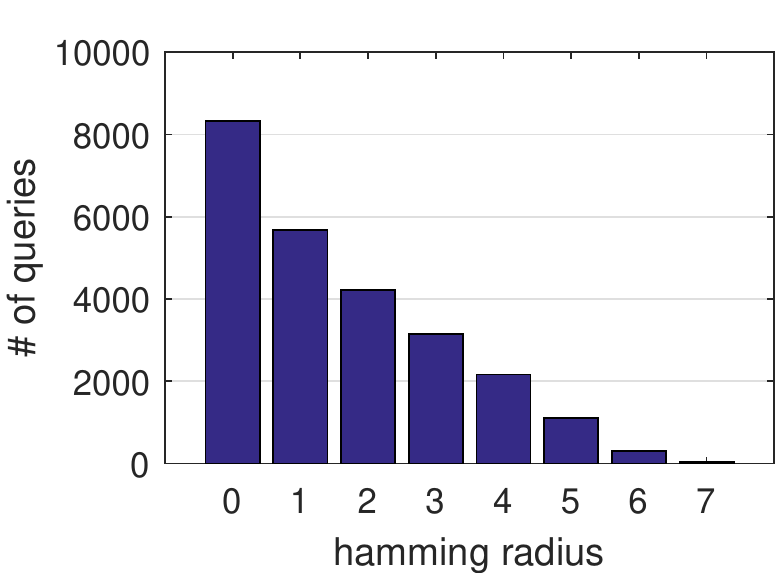}}
	\subfigure[34 bit]{\includegraphics[width=\FifthFigureWidth]{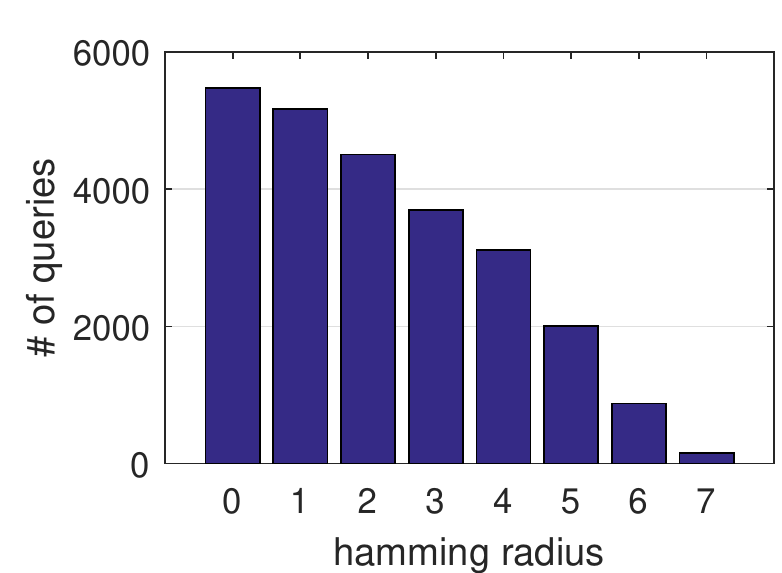}}
	\subfigure[36 bit]{\includegraphics[width=\FifthFigureWidth]{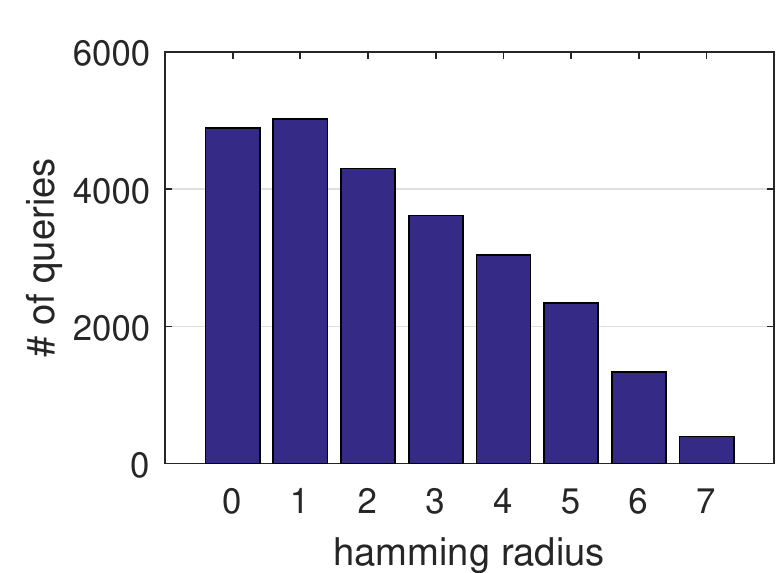}}\\
	\subfigure[16 bit]{\includegraphics[width=\FifthFigureWidth]{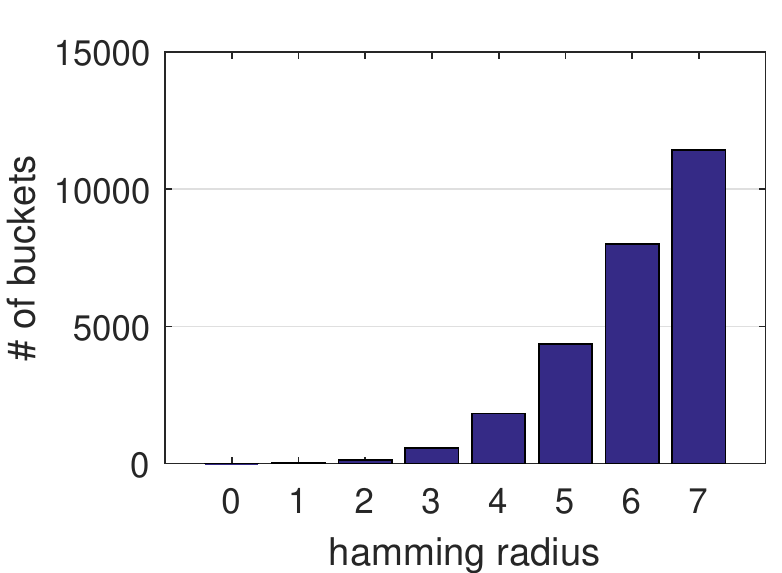}}
	\subfigure[24 bit]{\includegraphics[width=\FifthFigureWidth]{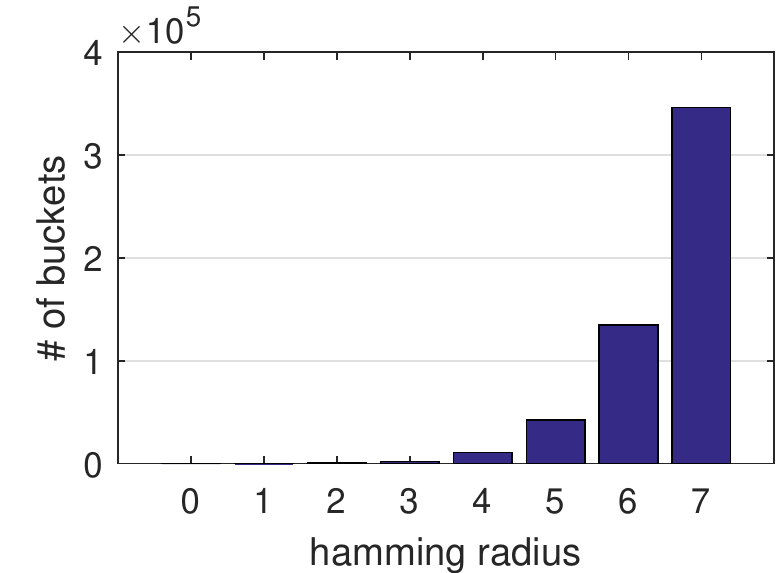}}
	\subfigure[32 bit]{\includegraphics[width=\FifthFigureWidth]{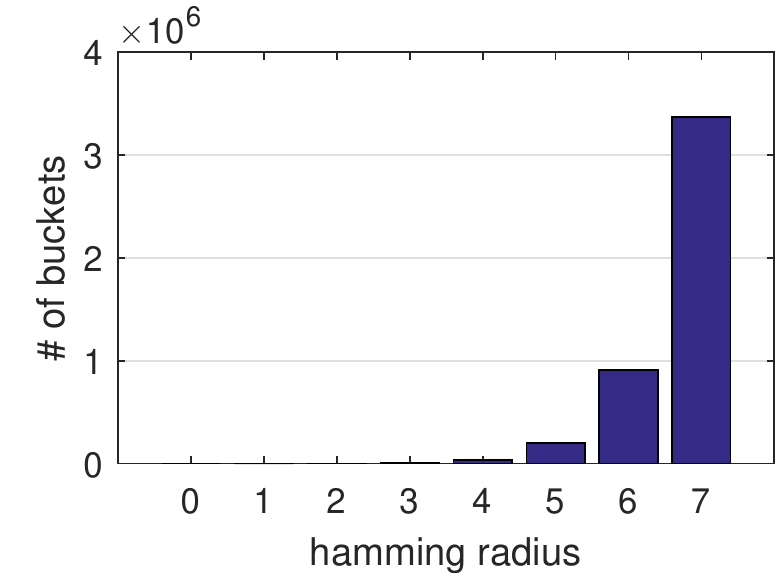}}
	\subfigure[34 bit]{\includegraphics[width=\FifthFigureWidth]{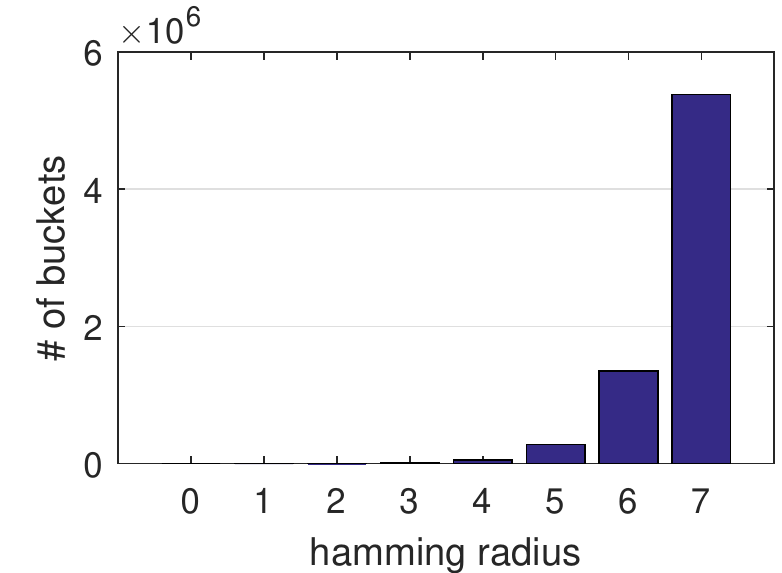}}
	\subfigure[36 bit]{\includegraphics[width=\FifthFigureWidth]{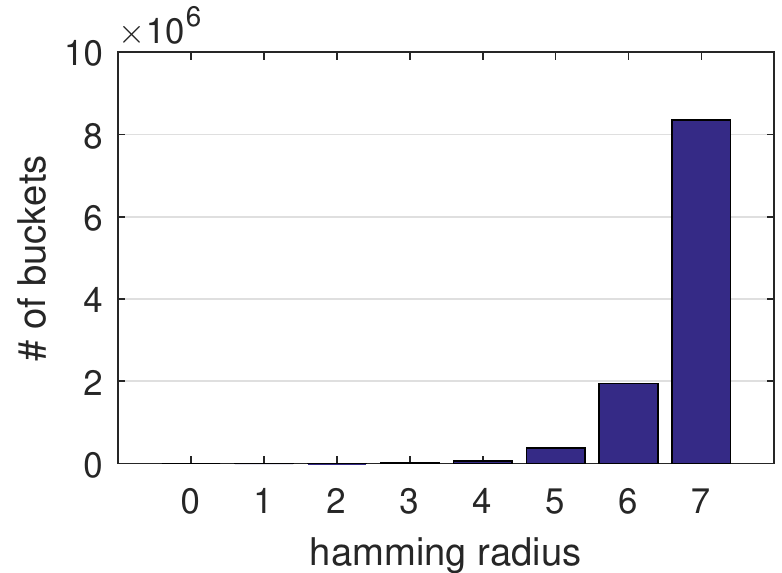}}
	\caption{The first row shows the number of queries (the total number is 25,000) which successfully located 100 images in different hamming radius of DLBH with different code length on partially supervised Imagenet. The second row shows that the number of hashing buckets grows quickly as the code length increases when the radius $r$ fixed. This figure explains why the locating time grows quickly as the code length increases.}
	\label{RadiusVSLen}
\end{figure*}

\subsubsection{Partially supervised setting} 

This time we use a more realistic setting by using only 10\% images in the base set as the supervised training image. More details of each set's size can be see on Table \ref{table:10p_imagenet_dataset_setting}. For the reproduction purpose, we release the learned 2048-dimensional features and the train/val/base/query splits\footnote{http://www.cad.zju.edu.cn/home/dengcai/Data/ANNS/ANNSData.html}.

Figure \ref{fig:res_ILSVRC15-P} shows the performance of all the compared hashing methods on the partially supervised Imagenet and all the hashing methods use 32-bit code. 

We can clearly see the advantage of using multiple table trick for LSH and IsoH. When we consider Precision@100, the DLBH method (the best performed deep hashing method among the three) does have some improvement over single table LSH and IsoH. However, 16 tables IsoH significantly outperforms DLBH. 

Overall, if the search time is correctly measured, the multiple tables trick is used and a large and complicated dataset is used, the claim in most of the deep hashing papers becomes wrong.

When a large scale and complected dataset is used, the precisions of all the hashing methods converge to the precision of the bruteforce search. It seems that the precision of the bruteforce search is the upper-bound for all the hashing methods. This never happened when a simple dataset is used or the label of all the base images are known. 

The real CBIR system has much larger size and much more complicated images. Our experimental results suggest a two-step approach for CBIR: 1) using complicated models (\eg, deep learning) to learn semantic real value features and 2) using advanced ANNS methods \cite{MalkovYHNSW16,NSG} to achieve fast retrieval.

\subsubsection{Effect of the code length}
Most of the existing deep hashing papers \cite{AAAI14SuperHashing, CVPR15DHCompactBinaryCode, CVPR15DeepSemantic, CVPR15CNNH, AAAI16DHNetwork, AAAI16DQNetwork, IJCAI16DPSH, IJCAI16DHEncoder, CVPR16DSH, ECCV16BinaryDNN, AAAI17DHJoint, AAAI17TransitiveDH, AAAI17PairwiseDH, IJCAI17DisDH, IJCAI17NolinearDH, IJCAI17LCDH} report the performances of various hashing methods increase as the code length increase. In this subsection, we will carefully re-examine this conclusion.

Figure \ref{fig:code_length} shows the performance of DLBH with different code length on the partially supervised Imagenet. Figure \ref{fig:code_length} (a) shows the {\em precision-time} curves while Figure \ref{fig:code_length} (b) shows the {\em precision-\# of retrieved samples} curves. From Figure \ref{fig:code_length} (a), we can see the best performance of DLBH is achieved when the code length is 32. As the code length further increases, the search time increases faster. Figure \ref{fig:code_length} (b) shows the performance of DLBH consistently increases as the code length increases. This simply because the time reported in Figure \ref{fig:code_length} (a) includes locating time and scanning time while the \# of retrieved samples in Figure \ref{fig:code_length} (b) can only reflect the scanning time. Ignoring the locating time is the reason that most of the deep hashing papers \cite{AAAI14SuperHashing, CVPR15DHCompactBinaryCode, CVPR15DeepSemantic, CVPR15CNNH, AAAI16DHNetwork, AAAI16DQNetwork, IJCAI16DPSH, IJCAI16DHEncoder, CVPR16DSH, ECCV16BinaryDNN, AAAI17DHJoint, AAAI17TransitiveDH, AAAI17PairwiseDH, IJCAI17DisDH, IJCAI17NolinearDH, IJCAI17LCDH} made a wrong conclusion.

From Figure \ref{fig:code_length} (b), we can see if \# of retrieved samples is set as 100 (the starting point of all the curves), DLBH (16 bits) reaches around 13\% precision while DLBH (36 bits) reaches almost 34\% precision. From Figure \ref{fig:code_length} (a), we can see DLBH (16 bits) uses around 7.5s while DLBH (36 bits) uses around 200s. Since \# of retrieved samples is fixed as 100, the scanning times for the two cases are the same. It is the locating time causing this 190s difference, \ie, the locating time (locating 100 samples) of DLBH (36 bits) is about 20 times than that of DLBH (16 bits). In other words, DLBH (36 bits) can find better candidates but needs longer time than DLBH (16 bits). 

Figure \ref{RadiusVSLen} provides an explanation. The first row of Figure \ref{RadiusVSLen} shows the number of queries (the total number is 25,000) which successfully located 100 samples in different hamming radius of  DLBH with different code length. Figure \ref{RadiusVSLen} (a) shows more than 18,000 queries retrieved 100 images successfully within hamming radius 0 (\ie, more than 18,000 queries only need to visit one hash bucket) when DLBH uses 16 bits. As a comparison, if DLBH uses 36 bits, Figure \ref{RadiusVSLen} (e) shows less than 6,000 queries successfully retrieved 100 images (\ie, the remaining queries need to visit more hash buckets). The second row of Figure \ref{RadiusVSLen} shows that the hash buckets number grows quickly as the code length increases when the radius $r$ fixed.

To retrieve 100 images, the number of buckets needed to be located increases almost exponentially as the code length increases. Thus, the locating time increases almost exponentially as the code length increases.

\section{Conclusion} \label{sec:conclusions}
Deep hashing methods recently attract a lot of research interest. The idea is attractive, but these methods' effectiveness and efficiency needs a carefully re-examination.  Three important factors are missed in most of the previous deep hashing papers. 1) They failed to correctly measure the search time. 2) They compared with the sub-optimal version of traditional hashing algorithms (failed to use the multiple tables trick). 3) They use some very small and simple data sets for evaluations. Under a more realistic setting, the results are quite surprising: several representative and state-of-the-art deep hashing methods cannot outperform the traditional multi-table IsoH.


\end{document}